# Design Considerations of an Unmanned Aerial Vehicle for Aerial Filming

D. Casazola [†1], F. Arnez [†2], H. Espinoza [*3]

[†]UAS Research Team - Fundación Jala, Bolivia
[*]TECNALIA, Spain

[1] dennis.casazola@fundacion-jala.org
[2] fabio.arnez@fundacion-jala.org
[3] huascar.espinoza@tecnalia.com

*Abstract*—Filming sport videos from an aerial view has always been a hard and an expensive task to achieve, especially in sports that require a wide open area for its normal development or the ones that put in danger human safety. Recently, a new solution arose for aerial filming based on the use of Unmanned Aerial Vehicles (UAVs), which is substantially cheaper than traditional aerial filming solutions that require conventional aircrafts like helicopters or complex structures for wide mobility. In this paper, we describe the design process followed for building a customized UAV suitable for sports aerial filming. The process includes the requirements definition, technical sizing and selection of mechanical, hardware and software technologies, as well as the whole integration and operation settings. One of the goals is to develop technologies allowing to build low cost UAVs and to manage them for a wide range of usage scenarios while achieving high levels of flexibility and automation. This work also shows some technical issues found during the development of the UAV as well as the solutions implemented.

K*eywords – UAV, Sports, Multirotor, Autopilot, Stabilization, Navigation, Maneuvers, Vision, Control, On-Board Computer.*

## I. INTRODUCTION

The conceptualization of Unmanned Aerial Systems (UAS), also commonly referred to as Unmanned Aerial Vehicles (UAV) or drones, brought a huge variety of civil applications ranging from surveillance and monitoring, emergency assistance (rescue, fire, disasters, etc.) or delivery/transport of services, since it offers a stable platform that can be obtained at affordable costs. With the advent of this technology, aerial filming took a huge leap forward as UAVs are an ideal tool to achieve shots previously deemed impossible and to meet the demands and aesthetics of applications such as sports filming.

As stated on [1] and [2], UAVs are air vehicles and associated equipment that do not carry a human operator, but instead fly autonomously or are remotely piloted. In an UAS, the UAV is merely part, albeit an important part, of a total system. The whole system comprises: (a) a control station, which includes the interfaces between the operators and the rest of the system, (b) the aircraft carrying the payload (sensors, cameras, etc.); and (c) the communication system between the control station and the aerial vehicle. As a matter of simplicity, we use "UAV" indistinctly in this paper for both the system and the vehicle concepts.

UAVs are suitable for aerial filming with a number of benefits compared to conventional methods. As pointed at [3], they can be used for high risk situations where mobility is a hard task (e.g. Dakar series is performed on rough terrains); their size permits to fly closer to an objective without impact to the environment (e.g. helicopters noise and wind generated); its deployment is faster (e.g. multirotors only need a few meters of open space); power consumption is reduced; and finally Global Positioning Systems (GPS) and Inertial Measurement Units (IMU) within Autopilots allows highly complex maneuvers and automated navigation with precise flight paths using the navigation and stabilization modules.

UAVs are complex systems with a full integration between software, hardware and aircraft mechanic systems. UAVs design must take into account several parameters like the mechanical parts, electrical parts and compatibility, control and processing modules, communication links, and finally all the accessories required for the specific application (e.g. a high resolution camera for aerial filming). UAV components also require an adequate interaction between all of them so the platform is considered as stable and controllable as required.

This paper describes design considerations for a small UAV and its components aimed at supporting sports aerial filming, particularly considering the distinctive operational conditions of Bolivia. The paper also presents an integration model between aircraft mechanics, hardware and software is proposed for the assembly of the UAV components, including any test activities during the development. Finally, we present a preliminary approach to automatically track an object that represents an element to be followed by the UAV vision and navigation system.

## II. RELATED WORK

Development of small UAVs is a novel problem. Most of information about parameters and available system components is based on a compilation of different projects, published in Internet, which have specific requirements and operations limitations.

With regard to UAV system integration, authors in [4] offer a detailed classification of different UAVs based on their size and capabilities at specific operational conditions, as well as information about research organizations and current trends. The work published in [5] presents a detailed component specification and module interconnection with focus on vision, navigation and control algorithms. Paper [6] discusses sensor and their influence on state estimation models and automated flight modes. A detailed component description is given at [7] with a highlight on features concerning fault tolerance and autonomy issues. On [8], authors show an assembly process and part selection for a quad-copter. Finally at [9], we can see a detailed design process and evaluation of each component with its impact on performance, along with architecture models for specific applications.

Control processing associated to autopilots and on-board computers are at the core of UAV systems. On [10], the authors provide a comparison between open source autopilots, their internal capabilities and control techniques with selection guidelines. At [11], specifications for development and selection autopilots is presented. The three publications referred at [12], [13] and [14] introduce the open source Pixhawk autopilot project, including its architecture and features explained with a vision navigation approach, as well as interconnection diagrams with components and connection, configuration and calibration issues. The work at [15] describes a specification for a Pixhawk system together with remote communication links for information transfer between the control systems of micro aerial vehicles based on MAVLink. Finally, [16] presents different middleware solutions for optimal resource sharing between different control modules.

With respect to operational conditions and application constraints, [17] explains specifications of complex scenarios where weather, altitude and distance parameters take a key role. Authors at [18] present an UAV for aerial filming, its methodology for system sizing, constraint evaluations and implementation debugging, as well as the system costs.

## III. SYSTEM REQUIREMENTS

The ultimate goal of the underlying work is to use a small UAV for the specific mission of tracking a moving object, in a sporting environment, with an onboard video camera. For autonomously performing such a mission, we must be able to robustly extract the object location from images despite difficult constraints such as object movements, image noise, illumination, and operational conditions. Aircraft systems have a strong dependency on operational conditions such as temperature, wind speed or altitude-air density, with which its aerodynamics can change drastically. The UAV under design is expected to run with the operational conditions of Bolivia, as extracted next from [19] and [20].

Our center of operations is on the city of Cochabamba with average operational conditions in the middle regarding the Bolivian extremes. Temperature at this region has an average of 17°C, oscillating between 9 to 27°C. Average wind speed can reach up to 12 km/h, however the annual average of 3.4 km/h. Average altitude near the operations center is around 2600 MASL where most of the tests were performed, although some tests took place at altitudes from 0 to 4000 MASL. With all this conditions air density is expected to be reduced about 20-40% compared to see level altitude. Table I provides a summary of these conditions.

TABLE I
OPERATIONAL CONDITIONS AT THE CITY OF COCHABAMBA

| PARAMETER | VALUE | UNITS |
| --- | --- | --- |
| Region | Sub-Andean | |
| Average Temperature | 17 | °C |
| Temperature Range | 9 - 27 | °C |
| Average Altitude | 2600 | MASL |
| Altitude Range | 0 – 4000 | MASL |
| Wind Speed Average | 3.4 | Km/h |
| Maximum Wind Speed | 12 | Km/h |

Conditions referred to the specific application are based on the requirements specified for sport aerial filming. These include: to have a closer look of the competition, to have a wide range of mobility, to be aware of power consumption and finally equipment capabilities (communication links, camera and stabilizer). Scenario conditions also delimit our system, operation is mostly on urban and suburban areas where buildings, trees and electrical wires/towers are the expected obstacles to avoid. With all these conditions manual flight of the system is restricted to a distance range up to 200 meters on horizontal direction and 100 meters on vertical direction, approach to the target is expected to be at least 10 meters away and 5 meters away from any static or dynamic obstacle. Figure 1 provides a sketch of the system interaction.

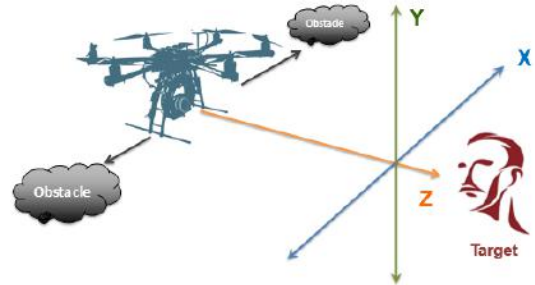

Figure 1. Sketch of the UAV interaction with its environment in order to perform the mission of video tracking.

## IV. SYSTEM SPECIFICATION

The underlying research work is based on the decision of using existing components to build the proposed UAV. Hence, the selection of the components and parts to meet the previously specified requirements is of core importance.

Figure 2 shows the UAV system architecture, which can be broken down into 5 subsystems:

### A. Mechanical Subsystem

The design of this subsystem considers parameters for the desired aerodynamic behavior, mission requirements and operational conditions.

#### 1) Air-frame

Air-frames define the aerodynamic structure, degrees of freedom and maneuvers of UAV systems. UAV air-frames can be of two main categories: fixed wing and rotary wing. Fixed wing UAVs have the advantage of being able to fly at high speeds for long duration with simpler structure. These UAVs have the disadvantage of requiring runway or launcher for takeoff-landing and not being able to hover. On the other hand, rotary wing UAVs or multi-rotors have the advantage of being able to hover, takeoff and land vertically (VTOL) with agile maneuvering capability at the expense of low speed and short flight time. The multi-rotor advantages are a key feature for sport filming without significant detriment of this mission.

Multi-rotors with more motors have a better redundancy, payload capacity and stability in detriment of complexity, power consumption, footprint (e.g. size) and, of course, their costs. We selected a hexacopter (six motors) since it is a low–cost solution suitable for aerial filming as they get good redundancy with acceptable stability and payload capacity.

#### 2) Propellers

Propellers are essential for UAV aerodynamics since they determine flight thrust and response rate. The main selection parameters are material, diameter, pitch (distance moved by the UAV with one revolution), and shape.

Carbon fiber propellers are expensive but highly efficient and resistant. Other materials are cheap but either fragile or inefficient (flexibility causes vibrations and power loss). Diameter and pitch adjust the thrust and response rate, but manufactures have specific sizes to avoid harms since engines work with bounded air-load. Diameter is maximized to get more thrust (5th order influence) resulting in low response rates. Shape is not considered meaningful as far as efficiency increase is not yet significant, and variable pitch profile is the most common shape.

A multi-rotor needs a balance of the different torque values generated by the rotors. To achieve this, a mixing of clock-wise (CW) and counter-clock-wise (CCW) propellers are often used.

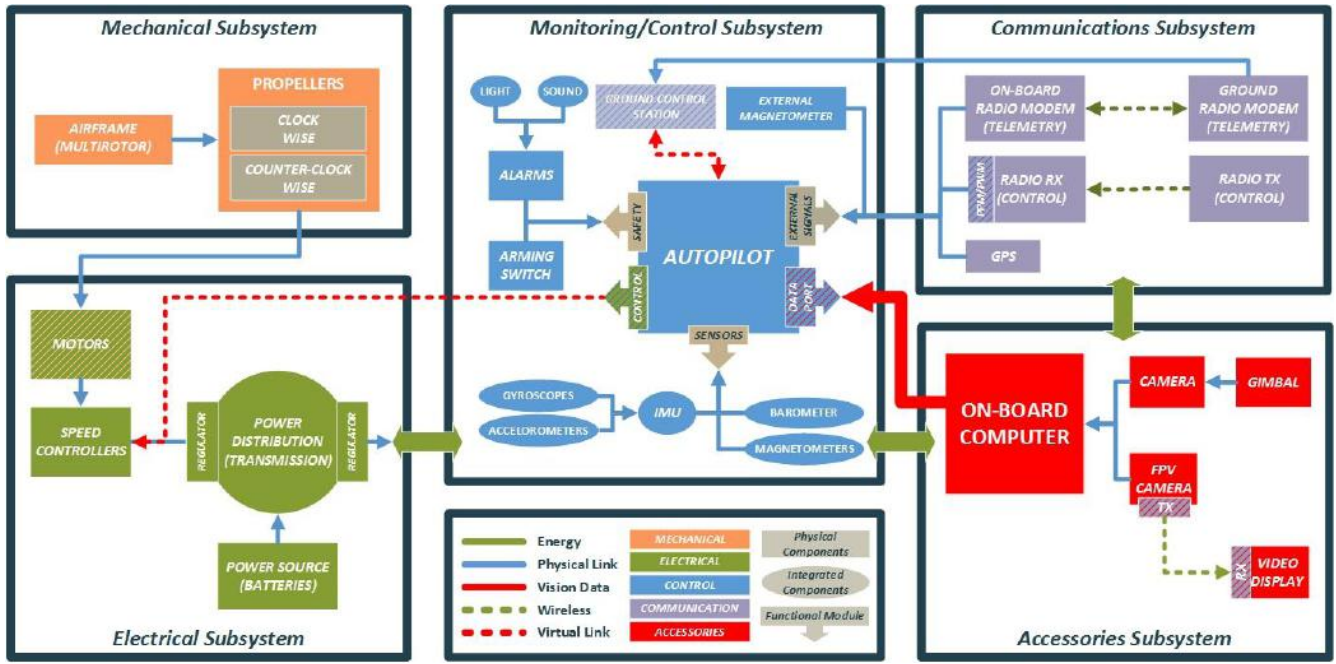

Figure 1. System architecture and subsystems composition.

### B. Electrical Subsystem

This subsystem addresses aspects of power source connectivity and compatibility with UAV electronic devices.

#### 1) Power Source (Batteries)

Options for UAV power source are electric batteries and fuel. The advantage of using batteries is that an electric system's thrust can be controlled more precisely and responds faster to throttle input. This is crucial to provide the differential thrust control required by this type of application. Also electric systems are more reliable, minimizing the possibility of crash due to motor shutdown or failure. The main drawback of electric systems is their flight time. Fuel (gas, nitrogen or gasoline) has high energy density and power; however, transmission mechanisms are hard to deal with. Fuel is more used for larger UAV systems (requiring more endurance). In our application case, aerial filming is considered a secondary video source regarding grown filming. This is why the video slots can be managed with short flight times (7-15 minutes) with the possibility of changing batteries in place.

Higher capacity and energy density (which means low weight) are key features for our application. LiPo batteries achieve this much better than other types. High discharge rate is important since VTOL demands and the number of cells define the operating voltage of the UAV system. Three or four cells are recommended with multi-rotor power demand to ensure compatibility with most of electronic components.

*1) Speed Controllers*

ESCs (Electronic Speed Controllers) vary an electric motor's speed and its direction according to the encoded speed (e.g., PWM to brushless signal). They are considered advanced devices with a built in firmware that sets cut-off voltage, braking and other features for optimal results. ESCs include Battery Eliminator Circuits (BEC) to energize low power devices (e.g. radio receiver). It must be ensured electrical compatibility with the motor as well as a proper calibration of PWM signals.

*2) Distribution Board*

A distribution board is essential for the electrical system since it interconnects all the electrical power signals between the power source and the components including stages that adjust voltage current and ratings to specific values.

C. Control Subsystem

These subsystem components continuously evaluate the UAV state and perform navigation and control functions.

*1) Autopilots*

The autopilot is considered the brain of the UAV system as it has direct control over its behavior. This board is responsible for the UAV stabilization and navigation, in both manual and automatic flight modes. It accepts commands and sensor data, and adjusts flight controls accordingly.

Proprietary autopilots are ready-to-go systems optimized for specific platforms with some tunable parameters. On the other hand, open-source-projects (OSPs) release blueprints, schematics and/or source code under an open-source model. Using open autopilots decreases costs dramatically and allows UAV solution developers to exchange hardware and software technologies. This release model also allows us to modify the core UAV software for our application, since we need to integrate vision and control functionalities to perform automatic tracking of elements in a sporting scenario.

OSPs capabilities are similar to advanced flight systems. They differ on failure treatment, redundancy and hardware specifications. Several OSPs exist but we selected the Pixhawk autopilot since it covers most of other OSPs features and they offer a unique relevant feature for us, which is a partial integration of vision systems, setting the base for vision based navigation. Other features include GPS-based waypoint navigation, camera stabilization, automatic flight modes and GCS (Ground Control Station) support.

Several modules are integrated within the autopilot. The Pixhawk architecture can be seen in figure 3. The sensor module has an IMU and consists of accelerometers that measure movement rate, and gyroscopes that measure rotational rate aligned to rotational axis (yaw, pitch and roll). Combining this sensor module with an estimator provides accurate measures of UAV position. Magnetometers are included for attitude estimation and a barometer for altitude estimation. A GPS is included to improve accuracy and to get the aircraft geo localization. The safety module establishes a safe operation under failure conditions. To this purpose, hardware and software are monitored to prevent, detect and manage failures. The control module is the processing unit that commands actuators and adjust parameters according to control algorithms and sensor inputs. The communication module defines the interface and protocol with external control and data signals.

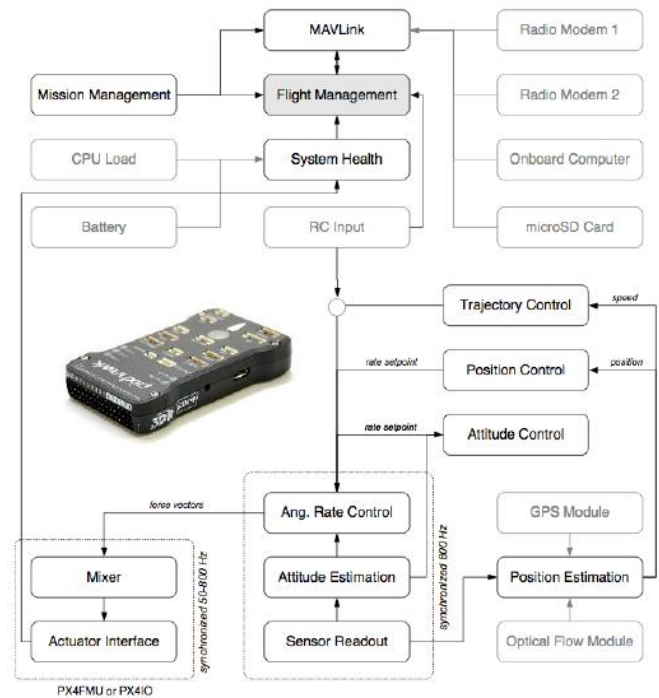

Figure 3. The Pixhawk architecture. Notice that black boxes are modules integrated in Pixhawk and gray ones are external.

*2) Ground Control Stations (GCS)*

A GCS is a land-based control center that provides the facilities for human control of UAVs. It allows monitoring of sensors and failsafe states, setting autopilot parameters and flight modes, and log history for analysis of system performance during missions. It is a critical module that manages automatic modes to prevent crashes and failures.

We decided to use MissionPlanner, a full-featured ground station application, which can be also used as a configuration utility or as a dynamic control supplement for UAVs. MissionPlanner along with telemetry hardware allows users to monitor UAV's status while in operation, record telemetry logs which contain much more information the on-board autopilot logs, view and analyze the telemetry logs, and operate your vehicle in FPV (first person view).

### D. Communication Subsystem

This subsystem manages the wireless links between the autopilot, GCS and other sources (e.g. GPS).

*1) Global Positioning System (GPS)*

The main GPS functionality is to estimate the global location of the UAV based on specialized satellite signals. GPS options are multiple but interconnection with the autopilot and accuracy are critical. Automatic flight modes are based on GPS data to correct positions and this limits the operation to outdoors and clear sky places where at least 9 satellites are detected (for an error of 1 meter). Indoor places does not permit this accuracy, since structures interfere satellite signals.

*2) Radio Control*

A radio transmitter, together with the interface for manual UAV piloting and the radio receiver, which receives PWM control signals for the autopilot, establish this link. Digital Spectrum Modulation (DSM) technology is suitable in civil applications, allowing the operation band within 2.4 GHz and the use of pure and mixed RC channels. Transmitter controls can be configured with specific functions and at the receiving end a PWM/PPM conversion takes places. The selection process considers nominal range and available channels.

*3) Telemetry*

Telemetry offers a feedback of the system state. It interconnects the GCS with the autopilot by using transceiver components. To avoid interference with other known systems, 450 or 900 MHz are used with a specialized protocol called MAVLink for data transfer. The range of these devices are in the order of kilometers and they can be used to bypass manual control through this link for greater distances.

### E. Accessories Subsystem

This subsystem changes drastically with the type of application (other subsystems might not change at all). In our case, the requirements for this subsystem are defined by the targeted application, i.e., aerial filming sports with a computer vision approach for objects/persons tracking.

*1) Camera*

The main resource for aerial filming is the video camera and its selection analysis is quite extensive. Specialized devices are available that fulfill different requirements of image quality for filming sports. High resolutions (1080p) can be achieved and special lens get a wide field of view (some have and enjoyable distortion). Illumination and focus are suited for outdoor environments and shockproof is tested for extreme sports conditions. Automatic zoom systems are not always available since they are not simplr to control. A critical condition is that images must be transferred on real time to a ground processing unit for object/human tracking and filming.

*2) Gimbal*

Vibration and movement inertia are critical for aerial filming. These unwanted effects produce camera instability and hence low video quality. A gimbal counteracts these movements with an IMU and position-controlled motors that adjust rotational changes of the camera axis (pan, tilt and yaw). While servomotors neutralizes slow but strong movements such as movement inertia, brushless motors counteract fast but weak movements such as vibrations. Gimbals with 2 degrees of freedom are common and simple, but 3-axis-gimbals are recommended if UAV movement is limited (e.g. more power consumption) or dynamics of filming demands. Some advanced control boards have a firmware to adjust parameters and features for a better performance.

*3) On-Board Computer*

This computing element is intended to process the vision algorithms. Computer vision algorithms manage resource demanding tasks, not necessarily complex but repetitive. This is also aggravated, and unfortunately this is aggravated with real-time requirements, like the approach with target following.

Workstations get good performance, as resources are easily adjusted, in contrast, on-board systems have limits with weight and energy consumption and to get acceptable performances, hardware specs need to get closer to workstations. Processing capacity depends on clock speed. Current CPUs manage about 1GHz. Processor architecture, either x86 or ARM are compatible with development tools. The number of processors and cores, along with associated scheduling mechanisms take advantage of parallel computing. Memory management, RAM and memory modules may limit response times. Another key point to be considered is compatibility with peripherals, including cameras and autopilot: Finally, we must consider power consumption, our main constraint, to get a sufficient UAV flight time.

*4) First Person View (FPV)*

Manual control at any required time of an UAV is a requirement due to safety reasons. Also, due to safety reasons, the pilot must have a visual line of sight (VLOS) of the UAV. However, this is not always possible. To replace this, a FPV has an on-board camera, with the front view of the UAV, transceivers and a display. It enables UAV perception similar to a pilot being on-board, which permits manual control without VLOS. It operates at 5.8GHz band and the most critical parameters are range, bandwidth and delay of the communication messages since it limits usefulness of the system for non-VLOS manual flights.

*5) System sizing*

Many different variables were evaluated throughout this section and merged them is time-consuming and complex. Fortunately, some tools have been designed and optimized, which use aerodynamics formulas, which allow us a fast evaluation of our requirements. Our approach is based on [21], which is an advanced tool that permits to tune component parameters and gives tips in order to achieve the desired system performance with an accuracy of $\pm 10\%$. The result of the UAV platform, with the specification considerations, can be seen in figure 4.

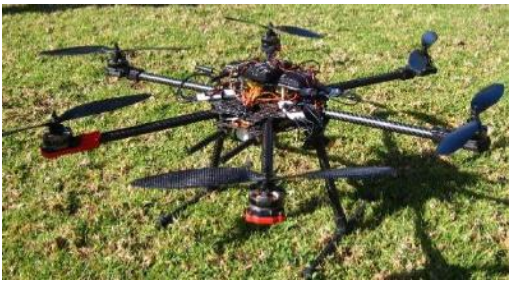

Figure 4. Hexacopter assembled for sports aerial filming. The targeted application is object/human tracking by using embedded, on-board computer vision.

## V. VISION AND CONTROL

In this section, tools and techniques used for control of the micro aerial vehicle (MAV) assembled at this work is presented for a vision navigation approach.

### A. MAVLINK Communication Protocol

The MAVLink open source communication protocol is designed to be used with many different wireless and wired links from serial (UART) to UDP. It is used throughout the system, not only for wireless communication with the operator control unit but it also serves as communication protocol between the flight computer [15] and the onboard main computer. It is able to use several links in parallel, allowing to use several redundant links, like long range XBee radio modems and 802.11n Wi-Fi and for even longer ranges it is possible to use GSM modems. The protocol uses only 8 bytes overhead per packet and has in-built packet drop detection. Due to the low overhead and efficient encoding it allows to execute the protocol on low performance microcontrollers [15]. MAVLink has been widely adopted in the open-source domain, and has been integrated into many different external micro aerial vehicle autopilots and different ground control stations as the main communication the protocol.

All MAVLink messages are generated from a XML protocol specification file. Currently code generators are available for C, Python and Java. The generated code contains all messages defined in the XML specification as well as functions for message packing and unpacking. This makes it very convenient to implement the protocol on different systems. Another important property of the protocol is its extensibility: Users can extend the default XML files and introduce their own custom messages. One key benefit is that the code gets automatically generated and there is no need to manually write the new code [15].

### B. Aerial Middleware

MAVCONN is a robotics toolkit tailored towards computer vision controlled autonomous lightweight MAVs. As we saw before, most components are connected via serial links and USB. The main problem with the existing middleware toolkits is that they do not scale down to this kind of links since every packet has to be transcoded by bridge processes, increasing the transmission delay. This new architecture design can be transparently used on different hardware links and minimizes the system complexity. Figure 5 shows a diagram for MAVCONN interconnection.

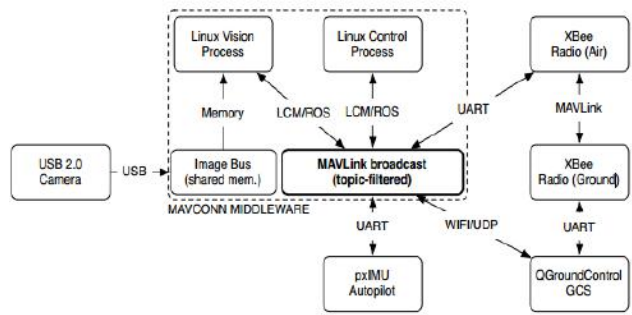

Figure 5. Typical MAVCONN network connection with their components and internal functional modules of the middleware.

The main communication layer uses Lightweight Communications and Marshalling (LCM) that outperforms Robot Operating System (ROS) in terms of latency [16]. An additional advantage is that LCM does not use any central server process for communication, directly increasing the robustness by eliminating the single point of failure when the central process crashes [16].

The MAVCONN middleware provides access to the different communication links to send MAVLink messages throughout the system and to off-board nodes. While MAVLink messages are used to send system states and control commands that are rather small, MAVCONN implements a convenient high-level application programming interface (API) to a shared memory based image hub on the onboard computer. It allows us to share the images of all cameras with an arbitrary number of Linux processes with the least overhead possible [15]. Although LCM is used as base middleware, it is possible to include ROS nodes into the system through the MAVCONN ROS-MAVLink bridge process that routes messages between both middlewares.

## C. Visual Control

A micro aerial vehicle, in this case an hexacopter, will not stay at a given position without continuous attitude and position control, as this type of MAVs are aerodynamically unstable. For this reason the autopilot calculates the desired attitude and controls the attitude using its onboard inertial sensor module. We can also take control of the MAV attitude by identifying objects of interest in the images provided by cameras. For this purpose the onboard computer gets images from a USB Camera and processes them to identify objects of interest. If an object of interest is identified, an image processing algorithm draws the center of mass of the object. Finally, the image coordinates of the center of mass are passed to the attitude control to track the desired object of interest. At figure 6 a schematic of the control process is presented.

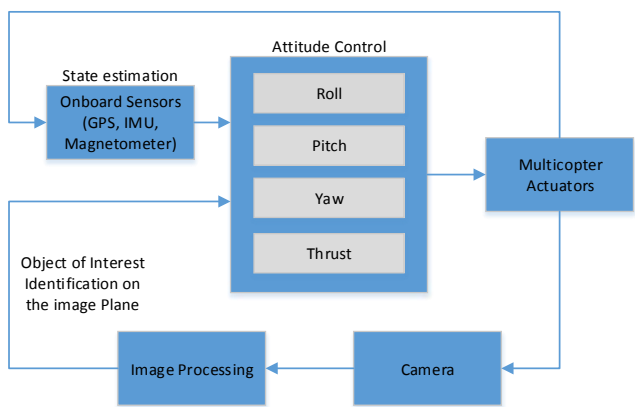

Figure 6. Control process with a computer vision approach. It can be seen that sensors provide the senses to the UAV for the control of specific parameters.

## VI. CONCLUSIONS AND FUTURE WORK

This paper describes the design considerations to develop a small UAV aimed at supporting sports aerial filming. We focused on the general system architecture, integration considerations (including mechanical, hardware and software components), communications, and basic software functionalities for control, navigation and vision. The specific operational conditions related to the geographical region of UAV deployment region have been also described.

Some of the main conclusions of the work are:

- We have developed a suitable platform for its deployment over Bolivia. It has freat performance of maneuverability, stability and navigation.
- The efficiency has been increase by 30% based on estimations.
- Components interaction was achieved by using an open middleware infrastructure.
- Visual tracking approach gives us high flexibility to film different kind of sports.

Our future work focuses on the improvement of the image processing algorithms, object avoidance implementation, and real tests deployment with safety considerations. On the other hand control process focus on Optimal Control techniques for attitude control.

### ACKNOWLEDGMENTS

The research leading to this paper has received funding from the Jala Group (Bolivia) and Tecnalia (Spain), under the Applied Research Programme - UAS Research project (hosted by Fundación Jala), The authors would also like to thank the others members of the UAS Research team, Alex Arenas, Héctor Quevedo, Jhonny Villarroel, Wilge Vargas, Willy Deheza, Luis Morales, Sergio Vargas, and Fernando Hinojosa, as well as the Jala management board.